\documentclass[10pt,letterpaper]{article}

\usepackage{cogsci}

\cogscifinalcopy 

\usepackage{pslatex}
\usepackage{xcolor}
\usepackage{apacite}
\usepackage{float} 

\usepackage{natbib}
\usepackage{graphicx}
\graphicspath{ {./images/} }
\usepackage{booktabs}
\usepackage{array}
\newcolumntype{M}[1]{>{\centering\arraybackslash}m{#1}}
\usepackage[export]{adjustbox}
\usepackage{subfigure}
\usepackage{subcaption}
\usepackage{multirow}
\usepackage{url}

\newcommand{\gtlogo}{\raisebox{0pt}{\includegraphics[scale=0.04]{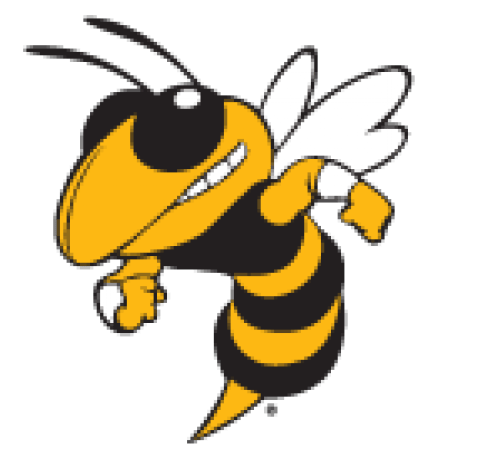}}}

\title{Incremental Comprehension of Garden-Path Sentences by Large Language Models: Semantic Interpretation, Syntactic Re-Analysis, and Attention}

 \author{{\large \bf Andrew Li, Xianle Feng, Siddhant Narang, Austin Peng, Tianle Cai, Raj Sanjay Shah, Sashank Varma} \\
  \{ali403, xianle.feng, snarang37, apeng39, tcai38, rajsanjayshah, varma\}@gatech.edu
  \\
   Georgia Institute of Technology \gtlogo
}

\begin{document}

\maketitle

\begin{abstract}

When reading temporarily ambiguous garden-path sentences, misinterpretations sometimes linger past the point of disambiguation. This phenomenon has traditionally been studied in psycholinguistic experiments using online measures such as reading times and offline measures such as comprehension questions. Here, we investigate the processing of garden-path sentences and the fate of lingering misinterpretations using four large language models (LLMs): GPT-2, LLaMA-2, Flan-T5, and RoBERTa. The overall goal is to evaluate whether humans and LLMs are aligned in their processing of garden-path sentences and in the lingering misinterpretations past the point of disambiguation, especially when extra-syntactic information (e.g., a comma delimiting a clause boundary) is present to guide processing. We address this goal using 24 garden-path sentences that have optional transitive and reflexive verbs leading to temporary ambiguities. For each sentence, there are a pair of comprehension questions corresponding to the misinterpretation and the correct interpretation. In three experiments, we (1) measure the dynamic semantic interpretations of LLMs using the question-answering task; (2) track whether these models shift their implicit parse tree at the point of disambiguation (or by the end of the sentence); and (3) visualize the model components that attend to disambiguating information when processing the question probes. These experiments show promising alignment between humans and LLMs in the processing of garden-path sentences, especially when extra-syntactic information is available to guide processing.

\end{abstract}

\section{Introduction}

Language is rife with ambiguity. Investigating how the human sentence parser handles this ambiguity has been important for revealing its processes, representations, and memory capacities. Of particular interest are \emph{garden-path sentences}, which are sentences that are temporarily ambiguous between two structural interpretations. Readers often choose the incorrect interpretation, for example, the one that is statistically more frequent in the linguistic environment \cite{macdonald1994lexical}. When they later reach the point of disambiguation, they are surprised and must then reanalyze the sentence to construct the correct parse tree and semantic interpretation. Surprisingly, the misinterpretation can linger and still be active at the end of the sentence \citep{christianson2001thematic, patson2009lingering}.

Large Language Models (LLMs) are deep neural networks trained on large corpora and range from multi-million parameter models like BERT \citep{devlin2019bert} and GPT-2 \citep{radford2019language} to state-of-the-art multi-billion parameter models like LLaMA-2 \citep{touvron2023llama} and GPT-4 \citep{achiam2023gpt}.
They have become more capable and have been claimed to reach human-level performance in general cognitive domains such as decision-making and problem-solving \citep{NEURIPS2020_1457c0d6, stiennon2022learning} and also on language tasks suhc as reading comprehension, grammar processing, and inference \citep{ye2023comprehensive, koubaa2023gpt}.
 
\begin{figure}
    \centering
    \includegraphics[width=0.43\textwidth]{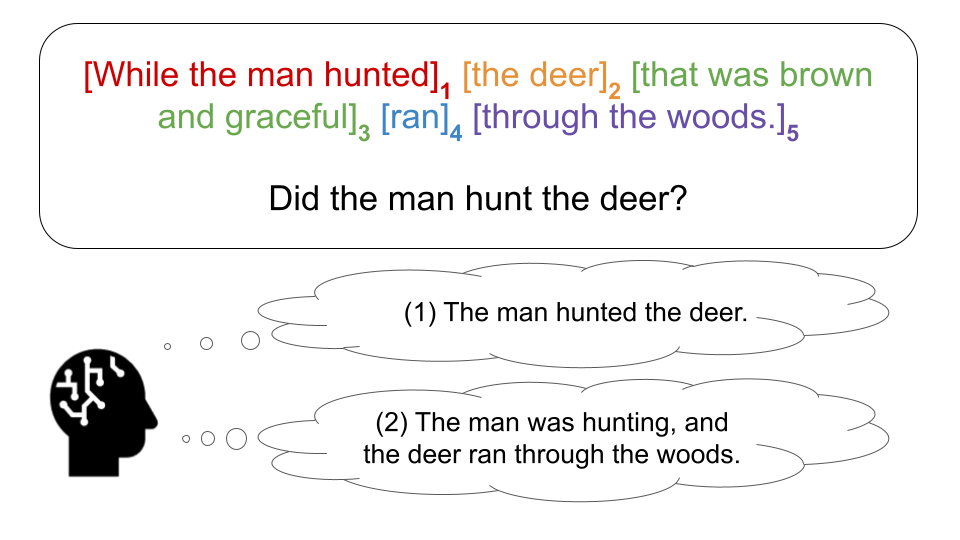}
    \caption{An example demonstrating the garden-path effect. During the incremental processing of this sentence, readers initially expect \textit{deer} to be the object of \textit{hunted}. Upon reaching the second verb \textit{run}, they realize that \textit{deer} is actually the subject of the second clause. This prompts a reanalysis of the sentence to the correct interpretation. However, the misinterpretation sometimes remains active even after reanalysis, and people still verify that ``the man hunted the deer''.} 
    \label{fig:problem_example}
    \vspace{-3mm}
\end{figure}
These successes have raised the question of whether LLMs are more than just engineering successes -- whether they are also viable \emph{scientific} models of human cognition. To support this claim, their performance must be measured and their representations examined for alignment to human behavioral signatures \citep{ivanova2023running, shah2023numeric, kallini2024mission, bhardwaj2024pretraining, cog_sci_typicallity}. Only then can their correspondence to human cognition be properly evaluated. Here, we do so for the processing of temporarily ambiguous sentences. We move beyond standard surprisal-based measures \citep{hale-2001-probabilistic, LEVY20081126, wilcox-etal:2020-on-the-predictive-power} and directly probe the semantic interpretations, implicit syntactic parse trees, and attention mechanisms as transformer-based LLMs incrementally process garden-path sentences.

\subsection{Garden-Path Effects}
Garden-path phenomena have long been studied in psycholinguistics. Here, we focus on two studies of the phenomenon exemplified in Figure \ref{fig:problem_example}. \citet{christianson2001thematic} had participants read 24 such garden-path sentences and tested their final understanding using two yes/no comprehension questions. For the sentence shown in Figure \ref{fig:problem_example}, the questions were:

       (1) Did the man hunt the deer?

       (2) Did the deer run through the woods?

\noindent Interpretation (1) is consistent with the transitive verb interpretation of the sentence, which is assumed during the ambiguous region (i.e., chunk 3) but ultimately proves to be incorrect. In particular, it is inconsistent with the occurrence of the second verb \emph{ran} in chunk 4, which is the point of disambiguation. And yet this incorrect interpretation lingers: Participants incorrectly verified (1) (i.e., responded 'yes')  about 60\% of the time. That said, they also computed the correct interpretation, verifying (2) nearly 90\% of the time. 

Additionally, \citet{christianson2001thematic} studied the effect of adding extra-syntactic information -- a comma between chunks 1 and 2 (i.e., "While the man hunted, the deer ran through the woods.") -- to signal the correct interpretation. This successfully minimized the ambiguity: a greater percentage of participants correctly rejected (1), the probe consistent with the misinterpretation, when the comma was present versus absent. \citet{patson2009lingering} found a similar result in their paraphrasing experiment.

\subsection{LLMs and Psycholinguistics}

A number of studies have investigated the alignment of LLMs with human sentence processing \citep{marvin-linzen-2018-targeted, wilcox-2019-cogsci}. 
Some of these studies, like the current study, have considered the processing of garden-path sentences \citep{jurayj-etal-2022-garden, wilcox-etal-2021-targeted, Wilcox2023}. 
However, most have focused on predicting word-by-word reading times, the coin of the realm in psycholinguistics, using \emph{surprisal} values derived from LLMs. Thus, they have shed little light on the research questions which animate the current study, which concern direct measurement of the semantic and syntactic interpretations that models form when processing garden-path sentences, and whether these interpretations shift following the point of disambiguation.


NLP researchers are becoming increasingly aware of the importance of representing ambiguity in LLMs, which is currently a challenge \citep{liu-etal-2023-afraid}. However, the human alignment of the technical solutions these researchers are developing is outside the scope of the current study.

\subsection{The Current Study}
	
The current study compared humans and LLMs on the incremental processing of garden-path sentences. We probed the online processing and final interpretations of a range of models to address the following research questions:

\begin{enumerate}
    \setlength\itemsep{0em}
    \setlength\parskip{0em}
    \setlength\parsep{0em}
    \item Do LLMs represent the \emph{semantic} misinterpretation of a garden-path sentence during the ambiguous region (i.e., chunk 3), and after they reach the point of disambiguation (i.e., chunk 4), do they switch to the correct interpretation?
    \item Is the switch from the misinterpretation to the correct interpretation at the point of disambiguation also reflected in the implicit \emph{syntactic} parse trees that LLMs construct?
    \item Is the attention mechanism of transformer-based LLMs sensitive to the point of disambiguation?
\end{enumerate}


We address these research questions using three novel methods that capitalize on the fact that LLMs can be directly interrogated in ways that human minds cannot. For the first research question, we present garden-path sentences to LLMs chunk by chunk. After each chunk, we used the comprehension questions -- examples (1) and (2) above -- to probe the strength of the misinterpretation and correct interpretation, respectively. This is a direct comparison between LLMs and humans that does not require indirect measures like \emph{surprisal} \citep{hale-2001-probabilistic, LEVY20081126, wilcox-etal:2020-on-the-predictive-power}.

A parse tree is a hierarchical representation of the syntactic structure of a sentence.  To address the second research question, we use the technique developed by \citet{manning2020emergent} 
to extract the parse tree at each chunk as an LLM incrementally processes a garden-path sentence. We evaluate whether this structure shifts at the point of disambiguation (i.e., chunk 4) from the misinterpretation to the correct interpretation.

The third research question is more exploratory in nature. We examine the attention weights of LLMs for evidence of sensitivity to the point of disambiguation (e.g., \emph{ran} in the example in Figure \ref{fig:problem_example}). These reflect how LLMs weigh the different elements of the input. \citet{vig2019multiscale} and \citet{clark2019does} introduced tools to visualize these weights in transformer models.
A substantial amount of linguistic information can be found in the attention weights of models \citep{clark2019does, liu2019roberta}. We ask whether this includes information about the point of disambiguation. 

\section{Method}

\subsection{Large Language Models}

Our study evaluated four LLMs ranging in performance, size, and architecture. 
We tested GPT-2 \citep{radford2019language} and LLaMA-2 \citep{touvron2023llama}, both decoder-only models with the latter being instruction-tuned. The third model was Flan-T5 \citep{chung2022scaling}, an encoder-decoder LLM. Finally, we evaluated RoBERTa \citep{liu2019roberta} as the most performant encoder-only-architecture model. RoBERTa is trained using a different paradigm (masked language modeling) than GPT-2 (next-token prediction) and LLaMA-2 and Flan-T5 (next-token prediction and instruction tuning).

 
\subsection{Tasks and Datasets}


We used two tasks originating in \citet{christianson2001thematic}. The first used garden-path sentences and yes/no comprehension questions like examples (1) and (2) above. Preliminary testing revealed that all models correctly answered probe question (2) corresponding to the correct interpretation. We therefore focused on the models' endorsement of probe question (1) corresponding to the misinterpretation.

To simulate the incremental processing, we presented sentences to models in chunks and interrogated their unfolding state. Specifically, we split sentences at the following points: (1) through the initial verb; (2) the misinterpreted ``direct object''; (3) the descriptive clause; (4) the second verb, or point of disambiguation where we expect models to reanalyze the semantics and syntax of the sentence; and (5) the rest of the sentence, where reanalysis might spill over to; see Figure \ref{fig:problem_example}.
We prompted the models as follows: prompt completion for GPT-2 and Flan-T5; in a chat format for LLaMA; and masked token prediction for RoBERTa. In all cases, we follow a question-answering template, prompting the language model with the garden-path sentence as context, the corresponding question, and ``Answer: ''. 

 
The second task spans the same 24 garden-path sentences and yes/no questions as the first task, but with a comma inserted after chunk 1 (i.e., the first verb; \emph{hunted} in Figure \ref{fig:problem_example}). This extra-syntactic information rules out the incorrect transitive interpretation of the first verb, and thus the misinterpretation that the noun phrase in chunk 2 (e.g., \emph{the deer}) is its direct object, potentially disambiguating the sentence.

With respect to the datasets, one is from (the final) Experiment 3B of \citep{christianson2001thematic}, where the performance measure was accuracy on the comprehension question (1) corresponding to the misinterpretation. The second is from \citep{patson2009lingering} who replicated this study but used a different accuracy measure: alignment of sentence paraphrases to the misinterpretation. Both datasets showed the same pattern of results: Inaccurate comprehension in the first task, but a positive effect of the extra-syntactic information and more veridical comprehension in the second task.


\subsection{Experiments and Measures}

\subsubsection{Surprisal}

Surprisal has been the dominant metric for linking
the processing of NLP models to psycholinguistic measures such as reading time \citep{hale-2001-probabilistic, LEVY20081126, wilcox-2020-real-time}. To establish a baseline and continuity with prior works, we also compute the surprisal values of the 4 models, averaged within each of the 5 chunks, for both tasks.

\subsubsection{Tracking Semantic Interpretations}

For probe (1) corresponding to the misinterpretation, we used token probabilities as an index of the likelihood of an LLM incorrectly answering ``yes''. That is, we collected the probability scores (logits) for the tokens ``yes" and ``no". We collected these measures after the processing of each of the five chunks.


We also computed the final answer accuracy for uniform comparison across models.
Specifically, after all five chunks had been processed, we tabulated which token -- ``yes'' corresponding to the misinterpretation and ``no'' to the correct interpretation -- had the higher probability. 

	
\subsubsection{Incrementally Extracting Parse Trees}

A parse tree represents the syntactic structure of a sentence. We extracted the incremental parse tree after processing each chunk. We first measured whether a model is misled during the ambiguous region (i.e., chunk 3; \emph{the deer that was brown and graceful} in Figure \ref{fig:problem_example}) and constructed the parse tree corresponding to the misinterpretation. We also measured whether, following the point of disambiguation (i.e., chunk 4; \emph{ran} in Figure \ref{fig:problem_example}), the model reanalyzes and constructs the parse tree corresponding to the correct interpretation.

We did so using the technique developed by \citet{manning2020emergent} to train and extract the parse tree embedded implicitly in the word vectors predicted by an LLM. This is done by training and applying a linear transformation on the embedding word vectors in the hidden layers, and thereafter constructing the minimum spanning tree (MST). In this MST, the distances of the words are the norms of the word vectors, and the dependencies of the words are the edges.

We trained the parse tree probe on GPT-2 and RoBERTa-large. (We did not have access to the computational resources to do so for Flan-T5 and LLaMA-2.) We extracted the parse tree after each chunk and looked for re-analysis following the point of disambiguation (i.e., chunk 4; \emph{ran} in Figure \ref{fig:problem_example}).


\subsubsection{Visualizing Attention Weights}

Attention is a key concept in transformer-based LLMs \cite{vaswani2017attention}. The self-attention heads of a model capture the inter-token relations given as input to the model as $(key, query, value)$ tuples. In BERT-like transformer architectures, there is a multi-head self-attention attached to each layer. This allows the model to capture a variety of structural relations within the token through a weighted dot product.

The attention heads can be visualized as heatmaps to promote explainability. \citet{clark2019does} and \citet{manning2020emergent} used this technique to find a correlation between different attention heads and various linguistic relations such as direct object and subject-verb agreement. We visualize the attention heads, specifically the attention between pairs of tokens, as heat maps. We focus on the attention weight (1) between the initial verb in chunk 1 and the misinterpreted ``direct object'' in chunk 2 and (2) between the misinterpreted ``direct object'' in chunk 2 and the second verb in chunk 4, for which it is the correct subject.
Thus, (1) represents the strength of the misinterpretation and (2) the strength of the correct interpretation. We then subtract (1) from (2) and interpret a positive value as evidence that the attention head is sensitive to the disambiguating information (and a negative value as evidence that it is not). We compute this value for all attention heads in all layers of the models.


\section{Results and Discussion}

\subsection{Surprisal Baseline}
\begin{figure}[h]
    \centering
    \includegraphics[width=0.38\textwidth]{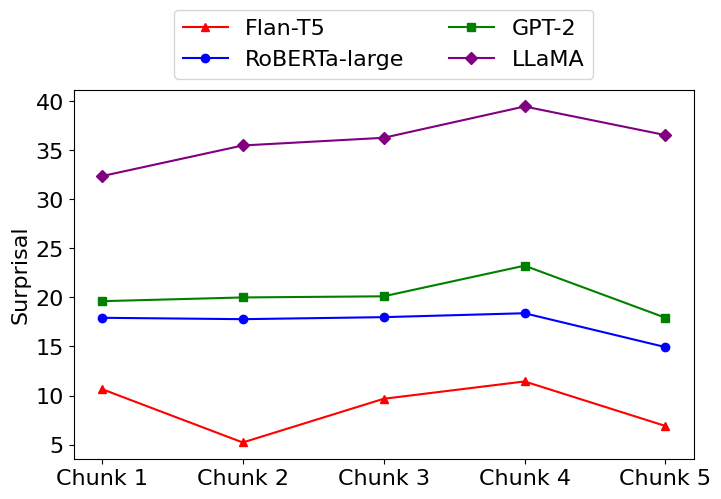}
    \caption{Surprisal in the first task (sentences with comma absent) across the four models.} 
    \label{fig:surprisal}
    \vspace{-3mm}
\end{figure}

\begin{figure}[h]
    \centering
    \includegraphics[width=0.38\textwidth]{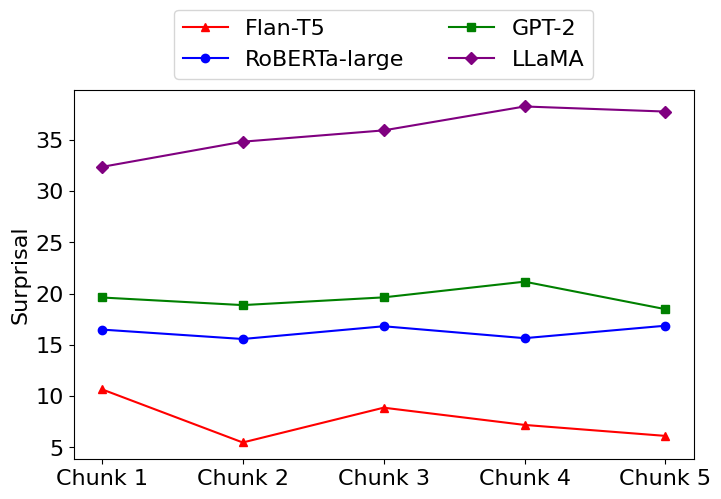}
    \caption{Surprisal in the second task (sentences with the disambiguating comma present) across the four models.} 
    \label{fig:surprisal_comma}
    \vspace{-3mm}
\end{figure}

In our baseline experiment, we compute surprisal for the four models on the first and second tasks, plotted in Figure \ref{fig:surprisal} and Figure \ref{fig:surprisal_comma}. For the first task, all models see a modest increase in surprisal at the point of disambiguation (chunk 4), as expected. For the second task, where the comma should disambiguate the sentence, RoBERTa and Flan-T5 show modest evidence of taking advantage of this extra-syntactic information and avoiding the garden path. By contrast, LLaMA and GPT-2 continue to show an increase in surprisal at chunk 4.

\subsection{Tracking Semantic Interpretations}
The first research question is when incrementally comprehending garden-path sentences, whether LLMs favor the misinterpretation during the ambiguous region (i.e., chunks 2 and 3) and then switch to the correct interpretation at the point of disambiguation (i.e., chunk 4) or afterward (i.e., chunk 5). 

\begin{figure}[h]
    \centering
    \includegraphics[width=0.49\textwidth]{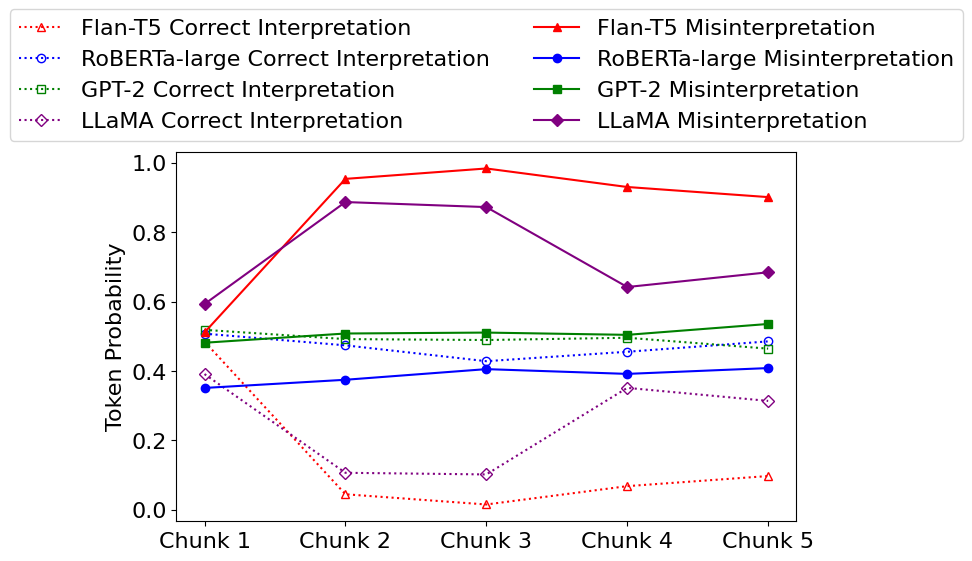}
    \caption{Semantic tracking of the mis- and correct interpretation in the first task (sentences with comma absent) across the four models. Critically, the probability of misinterpretation remains high even after the point of disambiguation.} 
    \label{fig:prob-average}
    \vspace{-3mm}
\end{figure}

Consider the first task, where commas are absent. Figure \ref{fig:prob-average} shows the probability of the misinterpretation (solid lines) and correct interpretation (dashed lines) across the five chunks for each of the four models. The first prediction is that the misinterpretation will be favored during the ambiguous region spanning chunks 2 and 3. This was the case for Flan-T5 and LLaMA, and by a smaller margin for GPT-2: the models assign a higher probability to ``yes'' to the verification probe (1), which is consistent with the misinterpretation, than to ``no''. The second prediction is that at the point of disambiguation, chunk 4, the pattern will reverse and the models will assign a higher probability to ``no''. This was \emph{not} the case; all models continued to endorse the misinterpretation. However, for LLaMA the probability of ``no'' decreased on chunks 4 and 5, which is a promising trend.

\begin{figure}
    \centering
    \includegraphics[width=0.49\textwidth]{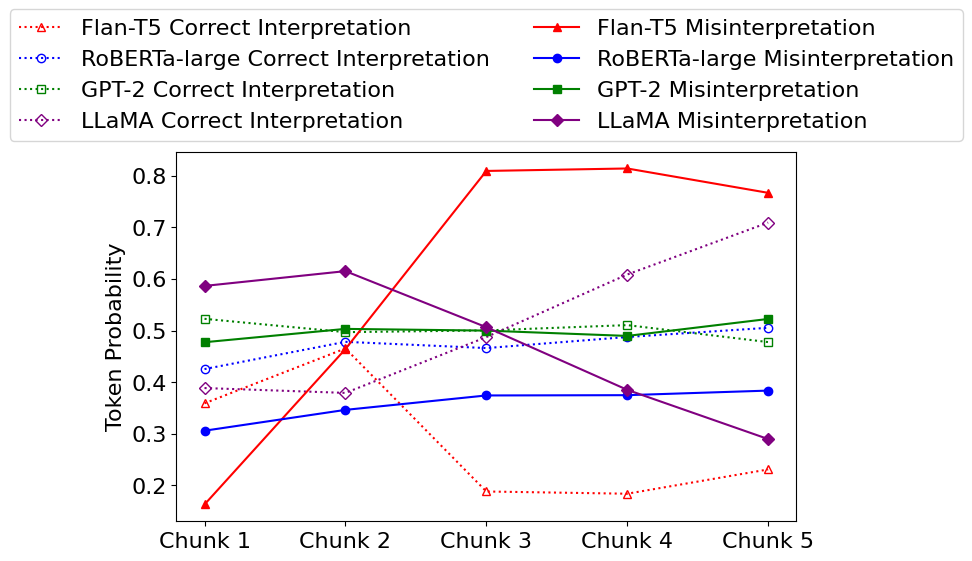}
    \caption{Semantic tracking of the mis- and correct interpretation in the second task (sentences with the disambiguating comma present). In the presence of this extra-syntactic information, the probability of misinterpretation decreases after the point of disambiguation for LLaMA, aligning with human performance.}
    \label{fig:prob-average-comma}
    \vspace{-3mm}
\end{figure}

By contrast, the models were more successful for the second task, where a disambiguating comma appears between chunks 1 and 2, suggesting that the noun phrase in chunk 2 should \emph{not} be misinterpreted as the direct object of the matrix verb in chunk 1; see Figure \ref{fig:prob-average-comma}. Specifically, LLaMA favors the misinterpretation during the ambiguous region (i.e., chunks 2 and 3). However, beginning at the point of disambiguation, the probability assigned to the misinterpretation decreases (a trend that continues in the final chunk of the sentence). This drop was statistically significant for LLaMA ($p = 0.003$) and also for GPT-2 (with $p = 0.05$). However, only LLaMA successfully switched to the correct semantic interpretation.


\begin{figure}[t]
    \centering
    \includegraphics[width=0.49\textwidth]{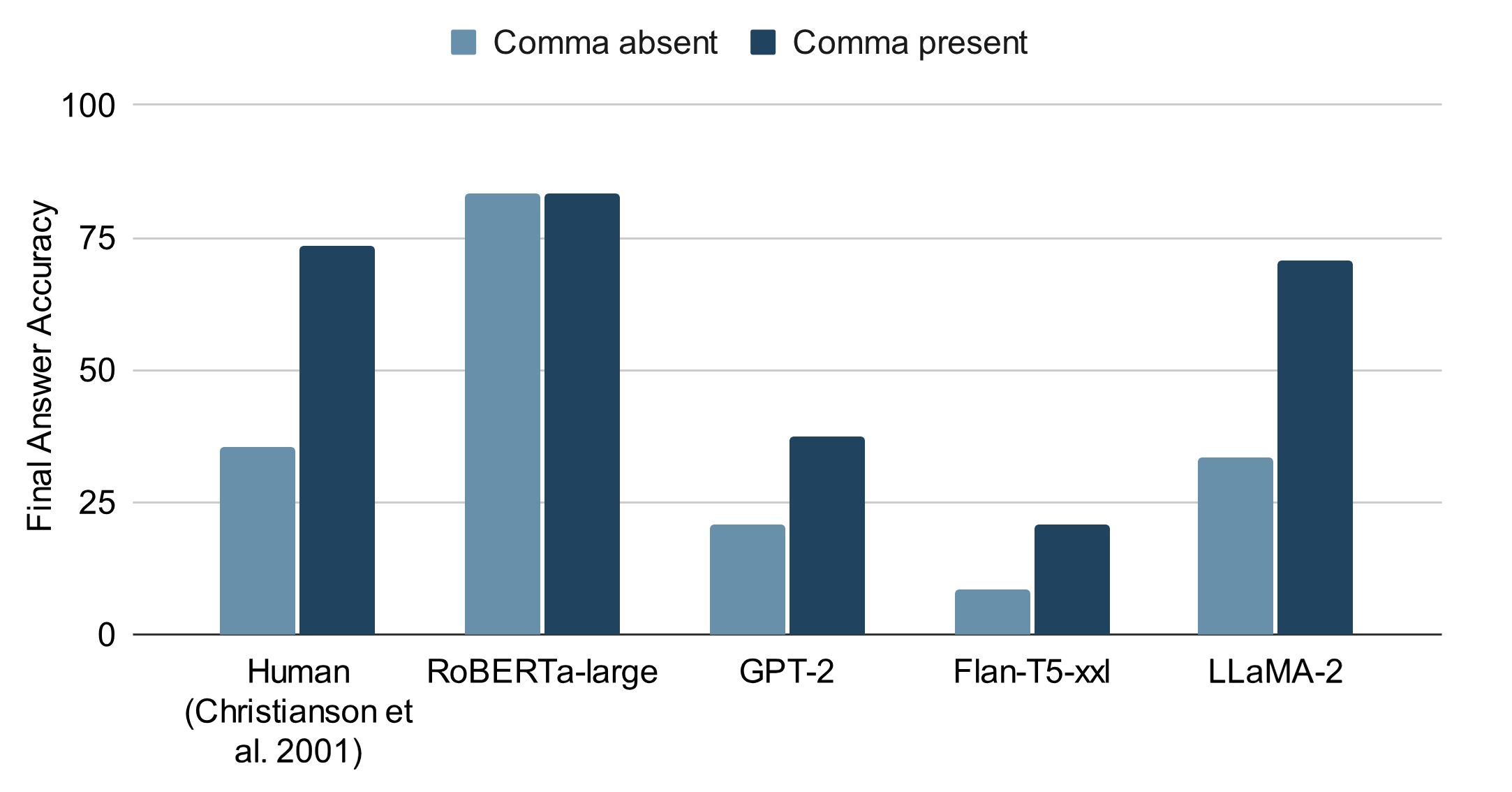}
    \caption{Final answer accuracy across models 
    }
    \label{fig:answer-accuracy}
    \vspace{-3mm}
\end{figure}

Turning from incremental semantic interpretation to the final semantic judgment, Figure \ref{fig:answer-accuracy} shows the percentage of garden-path sentences for which the probe question (1) corresponding to the misinterpretation was correctly rejected (i.e., had a ``yes'' response probability less than 50\%) at the end of the sentence for the two tasks and four models. Performance on the first task (i.e., sentences with comma absent) is shown in light blue, and on the second task (i.e., sentences with the comma present) in dark blue. Also shown is the human performance on the two tasks in the \citet{christianson2001thematic} study. LLaMA, GPT-2, and Flan-T5 show human-like performance in being garden-pathed for sentences without the comma but capitalizing on extra-syntactic information when it is present and recovering from the garden path. For all three of these models, accuracy is significantly greater on the comma-present vs. absent sentences ($p < 0.05$). LLaMA is notable in also showing human-like accuracies on the comma-present and comma-absent. By contrast, RoBERTa-large offers a poor fit for human performance.

\subsection{Incrementally Extracting Parse Trees}

\begin{figure}[h]
    \centering
    \hspace*{-0.12in}\includegraphics[width=0.51\textwidth]{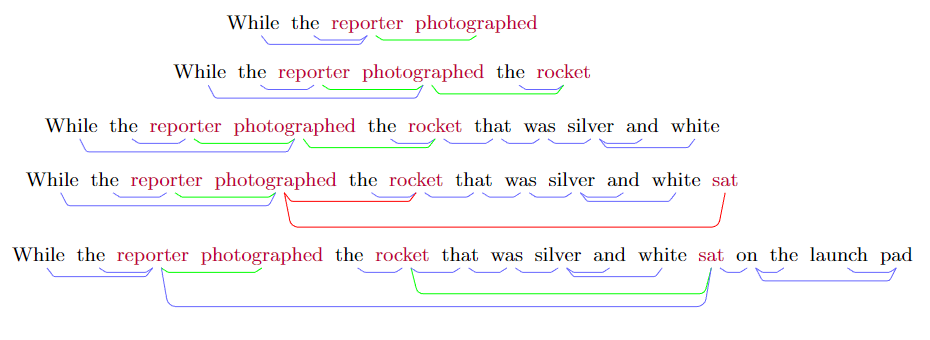}
    \caption{Example of incremental parse tree construction in RoBERTa-large.}
    \vspace{-3mm}
    \label{fig:parse-tree-construction}
\end{figure}

The second research question concerns syntactic reanalysis: when LLMs reach the disambiguation point of a garden-path sentence, do they reanalyze the implicit parse tree they are constructing and shift to one consistent with the correct interpretation? Figure \ref{fig:parse-tree-construction} shows the incremental parse trees across chunks 1-5 of the example sentence in Figure \ref{fig:problem_example}. These were extracted from the RoBERTa-large model. Note that the model initially incorrectly attaches \emph{the rocket} in chunk 2 as the direct object of the main verb \emph{photographed} in chunk 1. However, by the final chunk, it has shifted to the correct attachment, as the subject of the second verb \emph{sat}. (More generally, many of the extracted parse trees identify the correct dependencies of the words and subordinate phrases.)


\begin{table}[]
    \captionsetup{skip=2pt}
    \centering
    \caption{Proportion of sentences where the LLM makes the correct structural assignment shift for garden path sentences}
    \label{tab:parse-tree}
    \resizebox{0.4\textwidth}{!}{%
    \begin{tabular}{M{1.21in}M{0.36in}M{0.36in}M{0.36in}M{0.36in}}
        \toprule
        \multirow{2}{*}[-1.6ex]{LLM / Human} & \multicolumn{2}{c}{Comma Absent} & \multicolumn{2}{c}{Comma Present} \\\cline{2-5}
        & Chunk 1-4 & Chunk 1-5 & Chunk 1-4 & Chunk 1-5\\
        \midrule
        GPT-2 & $12.50$ & $16.67$ & $45.83$ & $50.00$ \\
        RoBERTa-large & $29.17$ & $45.83$ & $50.00$ & $62.50$ \\
        \midrule
        \cite{christianson2001thematic} & - & $35.40$ & - & $73.40$ \\
        \cite{patson2009lingering} & - & $21.00$ & - & $62.00$ \\
        \bottomrule
    \end{tabular}
    \vspace{-3mm}
    }
\end{table}

The performance of GPT-2 and RoBERTa-large across the 24 garden-path sentences is summarized in Table \ref{tab:parse-tree}. The table also contains the human data from the \citet{christianson2001thematic} and \citet{patson2009lingering} studies. First, consider the model and human performance on the first task, where the comma is absent. Following chunk 4, the point of disambiguation, especially RoBERTa-large begins to shift to the correct parse tree. By the end of the sentence (i.e., chunk 5), it has computed the correct parse tree for 45.83\% of the sentences. This is comparable to the performance observed in especially the original \citet{christianson2001thematic} study. These findings coincide with those of \citet{slattery2013lingering}, who found that reinterpretation of garden-path sentences spills over to the few words after the point of disambiguation.

Consider the second task, where sentences contain a disambiguating comma between chunks 1 and 2. As Table \ref{tab:parse-tree} shows, GPT-2 and RoBERTa-large shift to the correct parse tree after chunk 4, the point of disambiguation, for 45.83\% and 50\% of the sentences, respectively. The percentages increase to 50\% and 62.5\%, respectively, by the end of the sentence (i.e., chunk 5). Again, the accuracy of RoBERTa-large is comparable to the human participants in both studies. To summarize, RoBERTa-large performs most similarly to humans.


We confirmed that these descriptive patterns are also statistically significant by $t$-tests. For GPT-2, the presence of a comma had a significant impact on the proportion of sentences on which the model switched to the correct parse tree at the point of disambiguation, i.e., after chunks 1-4 ($p<0.01$), as well as at the end of chunk 5 ($p<0.01$). The results were similar for RoBERTa-large: the presence of a comma had a significant impact on switching to the correct parse tree after chunks 1-4 ($p<0.1$). However, the percentages at the end of chunk 5 are comparable ($p>0.1$).  On the other hand, and collapsing across the two tasks, having seen chunk 5 had a significant impact on switching to the correct parse tree for RoBERTa-large ($p<0.01$). But it does not make much difference on the performance of GPT-2 ($p>0.1$).

\subsection{Visualizing Attention Weights}

The final research question is whether the attention mechanism of transformer models is sensitive to the point of disambiguation when processing garden-path sentences.
Because LLaMA-2 and RoBERTa-large showed the strongest alignment with human performance in tracking semantic interpretations and extracting parse trees, respectively, we focus on these models. 

We defined sensitivity as follows. LLaMA-2 is composed of 40 layers x 40 attention heads per layer; RoBERTa-large is composed of 24 layers x 16 attention heads. After a model processes a garden-path sentence, we can quantify the sensitivity of an attention head to the correct interpretation. Positive evidence is given by the attentional weight between the noun phrase in chunk 2 (e.g., \emph{deer}) and the verb in chunk 4 (e.g., \emph{ran}) representing the correct interpretation/attachment. Negative evidence is given by the attentional weight between the same noun phrase and the verb in chunk 1 (e.g., \emph{hunted}) representing the misinterpretation/incorrect attachment. Then, positive evidence minus the negative evidence gives us the desired index: more positive values indicate good sensitivity, and more negative values have poor sensitivity.

\begin{figure}[h]
\includegraphics[width=0.55\textwidth,center]{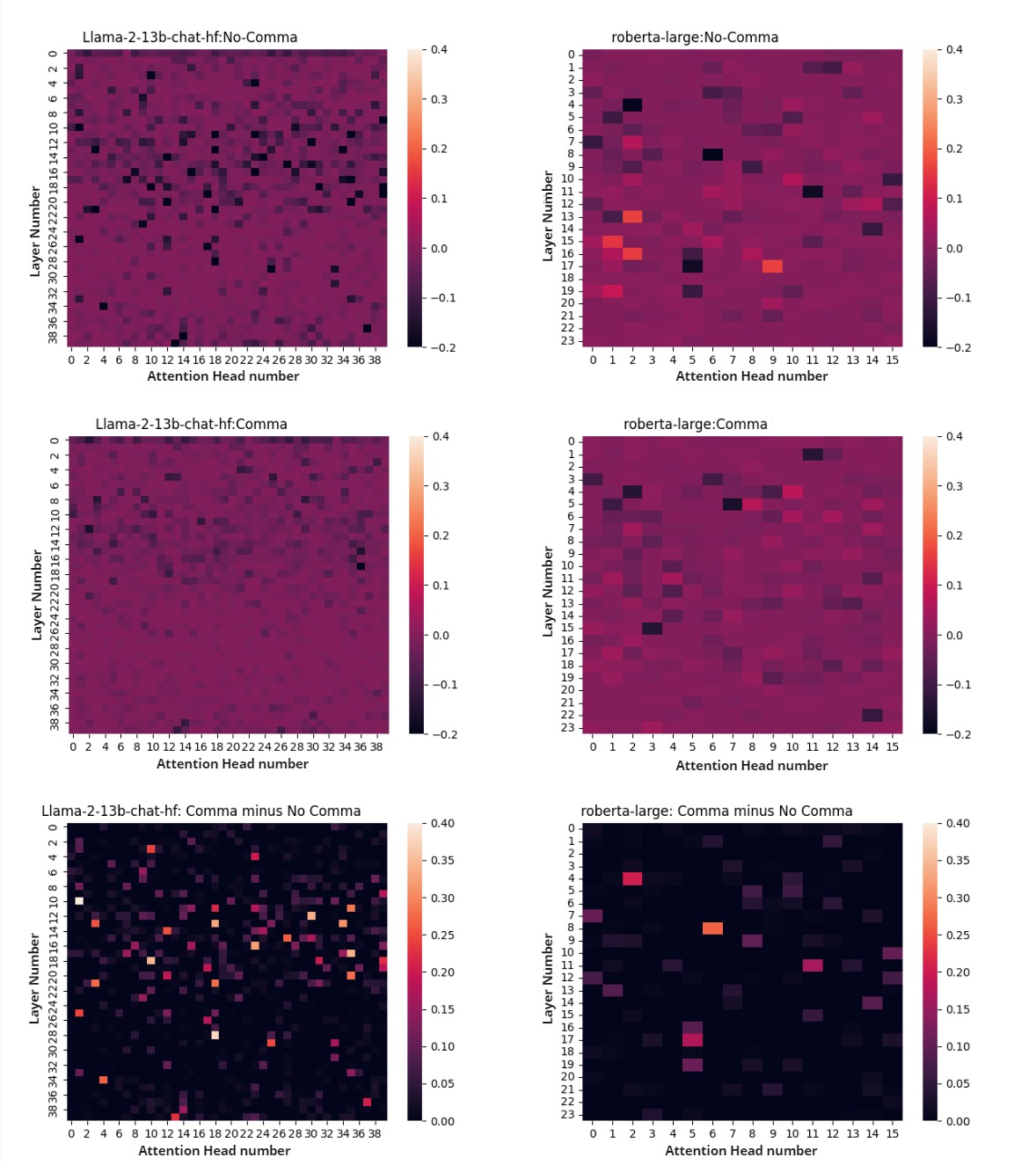}
    \caption{Sensitivity of the attention heads towards the correct interpretation in the presence of the comma.}
    \label{fig:attn-weights}
    \vspace{-3mm}
\end{figure}

Figure \ref{fig:attn-weights} shows the heatmap of the sensitivities across the attention heads of LLaMA-2 (left column) and RoBERTa-large (right column)
for the comma-absent sentences in the top row. The heatmaps for the comma present sentences are in the middle row, and the thresholded difference of the middle row minus the top row is in the bottom row. The bright cells in the top and middle rows indicate the attention heads sensitive to the point of disambiguation, and thus the correct interpretation/parse, for the sentences of the first and second tasks. The bright cells in the bottom row indicate the attention heads that show particularly increased sensitivity in the context of the disambiguating comma of the second task.


These findings are exploratory in nature, and thus valuable in the new questions they raise. For example, one is whether the same attention heads that are sensitive to disambiguating information for the garden-path sentences studied here are also sensitive to disambiguating information in other temporarily ambiguous sentence structures. 

\section{Discussion}

Cognitive scientists are increasingly investigating the value of LLMs as scientific models of the human sentence parser \citep{marvin-linzen-2018-targeted, wilcox-etal-2021-targeted, Wilcox2023}. The current study investigated the alignment between LLMs and humans in processing garden-path sentences. \cite{christianson2001thematic} and \cite{patson2009lingering} showed that when people process such sentences, misinterpretations can linger past the point of disambiguation, and people will verify probe questions consistent with them; see Figure \ref{fig:problem_example} for an example.
However, when the sentence is disambiguated early by adding a comma between chunks 1 and 2, people are less likely to do so.

We first showed that the classic surprisal metric shows some sensitivity to the point disambiguation. The first experiment, on semantic interpretation, found that RoBERTa, GPT-2, Flan-T5, and LLaMA-2 are garden-pathed on the first task, where the sentences have no commas, and misinterpretations lingered. However, on the second task, when the disambiguating comma was present, the larger models (i.e., GPT-2 and especially LLaMA-2) showed evidence of shifting to the correct interpretations. LLaMA-2, in particular, approximated human performance in its overall accuracy. The second experiment, on the incremental extraction of parse trees, found that GPT-2 and especially RoBERTa-large were sometimes able to switch to the correct parse tree at the point of disambiguation (i.e., chunk 4), and were even more successful afterward (i.e., chunk 5), especially when a comma was present. Here, RoBERTa-large approached human performance. The third experiment was more exploratory, showing that some of the attention heads of the LLaMA-2 and RoBERTa-large models are sensitive to the shift from the misinterpretation/incorrect attachment to the correct interpretation/attachment, as well as to the disambiguating information carried by the comma in the second task. Taken together, these results add to the growing evidence for LLMs as viable psycholinguistic models.

There are limits to our research that should be addressed in future studies. First, the four models we used span a range of architectures, training approaches, and sizes. However, they are far from exhaustive, and future work should evaluate a larger set of LLMs. Second, larger models tend to show emergent abilities \citep{wei2022emergent}. It is therefore important to run these experiments on larger models, both in upscaling the current models like LLaMA-70B and with newer LLMs like LLaMA-3 \citep{llama3modelcard} or Mixtral \citep{jiang2024mixtral}.


A final limitation is that the set of materials used in our experiments is small, containing only 24 garden-path sentences.
To increase the robustness and generality of our conclusions, it would be beneficial to repeat the experiments on a larger number of garden-path sentences for which human data is available, as well as on a broader range of temporary (syntactic) ambiguities.

\bibliographystyle{apacite}

\setlength{\bibleftmargin}{.125in}
\setlength{\bibindent}{-\bibleftmargin}

\bibliography{CogSci_Template}

\end{document}